\documentclass[11pt,a4paper]{article}

\usepackage[dvipsnames]{xcolor}
\usepackage[hyperref]{emnlp2018}
\usepackage{times}
\usepackage{latexsym}

\usepackage{url}

\usepackage{xr}
\usepackage{bm}
\usepackage{enumitem}
\usepackage{booktabs}
\usepackage{multirow}
\usepackage{graphicx}
\usepackage{amssymb}
\usepackage{amsmath}
\usepackage{verbatim}
\usepackage{breakcites}
\usepackage{pifont}
\usepackage{dsfont}

\newcommand{\bs}[1]{\textbf{\textsc{#1}}}
\newcommand{\s}[1]{\textsc{#1}}
\usepackage[hyphenbreaks]{breakurl}
\usepackage{url}

\usepackage{etoolbox}
\makeatletter
\patchcmd\@combinedblfloats{\box\@outputbox}{\unvbox\@outputbox}{}{%
   \errmessage{\noexpand\@combinedblfloats could not be patched}%
}%
 \makeatother

\aclfinalcopy 

\title{Picking Apart Story Salads}

\author{
Su Wang$^{1,2}$\ \ \ Eric Holgate$^{1}$\ \ \ Greg Durrett$^{3}$\ \ \ Katrin Erk$^{1}$
\\
$^{1}$Department of Linguistics \\
$^{2}$Department of Statistics and Data Science \\
$^{3}$Department of Computer Science 
 \\
The University of Texas at Austin \\
{\tt \small{shrekwang@utexas.edu}\ \  \small{holgate@utexas.edu}} \\
{\tt \small{gdurrett@cs.utexas.edu}\ \   \small{katrin.erk@mail.utexas.edu}}
}

\date{}

\begin{document}
\maketitle

\begin{abstract}
During natural disasters and conflicts, information about what
happened is often confusing, messy, and distributed across many
sources. We would like to be able to automatically identify relevant
information and assemble it into coherent narratives of what
happened. To make this task accessible to neural
models, we introduce \emph{Story Salads}, mixtures of
multiple documents that can be generated at scale. By exploiting the
Wikipedia hierarchy, we can generate salads that exhibit challenging
inference problems. Story salads give rise to a novel, challenging
clustering task, where the objective is to group sentences from the
same narratives. We demonstrate that simple bag-of-words similarity
clustering falls short on this task and that it is necessary to take
into account global context and coherence. 

\end{abstract}

\section{Introduction}
\label{sec:intro}

When a natural disaster strikes or a conflict arises, it is often hard to determine what happened. Information is messy and confusing, spread out over many messages, buried in irrelevant text, and even conflicting. For example, when flight MH-17 crashed in Ukraine in 2014, there were initially many theories of what happened, including a missile strike initiated by Russia-affiliated militants, a missile strike by the Ukrainian military, and a terrorist attack. There was no single coherent interpretation of what happened, but multiple, messy narratives, a \emph{story salad}. We would like to be able to automatically identify relevant information and assemble it into coherent narratives of what happened. This task is also the subject of an upcoming task at the Text Analysis Conference.\footnote{\url{https://tac.nist.gov/2018/SM-KBP/index.html}}

Picking apart a story salad is a hard task that could in principle make use of arbitrary amounts of inference. But it is also a task in which coherence judgments could play a large role, the simplest being topical coherence, but also narrative coherence~\citep{Chambers:08,Chambers:09,Pichotta:16, Mostafazadeh:17}, overall textual coherence~\citep{Barzilay:08,Logeswaran:18}, and  coherence in the description of entities. This makes it an attractive task for neural models. 

\begin{figure}[t]
\begin{center}
\scalebox{0.70}{
\begin{tabular}{@{}p{28em}}
\toprule
\textcolor{NavyBlue}{(A) Some of the prisoners were survivors of the Battle of Qala-i-Jangi in Mazar-i-Sharif.} 
\textcolor{NavyBlue}{ (A) Chechnya came under the influence of warlords.}
\textcolor{Sepia}{(B) The U.S. invaded Afghanistan the same year when several Taliban prisoners were shot.}
\textcolor{NavyBlue}{(A) Russian federal troops entered Chechnya and ended its independence.} 
\textcolor{NavyBlue}{(A) The Russian casualties included at least two commandos killed and 11 wounded.}
\textcolor{Sepia}{(B) The dead were buried in the same grave under the authority of Commander Kamal.}\\
\bottomrule
\end{tabular}}
\end{center}
\caption{A story salad involving two articles, about a Russian military operation in Chechnya \textcolor{NavyBlue}{(A)} and about a U.S. operation in Afghanistan \textcolor{Sepia}{(B)}. These two articles are topically similar but their mixture can still be disentangled based on narrative coherence.}
\label{tab-doc-mix}
\end{figure}

To make the task accessible to neural models, we propose a simple method for creating simulated story salad data at scale: we mix together sentences from different documents. Figure~\ref{tab-doc-mix} shows an example mixture of two articles from Wikipedia, one on the Russia-Chechnya conflict and one on a conflict between the U.S. and Afghanistan.  By controlling how similar the source documents are, we can flexibly adjust the difficulty of the task. In particular, as we show below, we can generate data that exhibits challenging inference problems by exploiting the Wikipedia category structure.\footnote{In particular, while we do not focus on creating mixtures with conflicting information, it can often be found in mixtures created based on Wikipedia categories, as shown in Figure~\ref{tab:wiki-hard-ex}.}  While this data is still simpler than story salads arising naturally, it approximates the task, is sufficiently challenging for modeling, and can be generated in large amounts.\footnote{The story salad data is available at {\url{http://www.katrinerk.com/home/software-and-data/picking-apart-story-salads-1}}. The Wikipedia salads are available for download directly and we have provided code to reconstruct the NYT salads from English Gigaword 5 (available as LDC2003T05).}


We explore some initial models for our Story Salad task. As the aim of the task is to group story pieces into stories, we start with straightforward clustering based on topic similarity. But topic similarity is clearly not enough to group the right pieces together. For example, the two articles in Figure~\ref{tab-doc-mix} are both about armed conflicts, but the Russia-Chechnya sentences in the example form a group \emph{in contrast to} the U.S.-Afghanistan sentences. To model this, we learn sentence embeddings adapted to the clustering task and with access to global information about the salad at hand. We also test an extension where to decide whether to group two sentences together, the model mutually attends to the sentences during encoding in order to better focus on the commonalities and differences of these two sentences. Both extensions lead to better models (6-13\% improvement in accuracy with a model incorporating both), confirming that the task requires more than just general topical similarity. But there is much room for improvement, in particular on salads generated to be more difficult, where performance is around 15 points lower than on arbitrary mixtures.

\section{Related Work}
\label{sec:related}

Building on early work in \emph{script learning} \cite{Schank:77}, \citet{Chambers:08} introduce \emph{narrative schema} and propose the ``narrative cloze'' task where the modeling objective is to predict the event happening next. The topic has since seen many extensions and variants coupled with increasingly sophisticated models \cite{Chambers:09} including neural networks \cite{Granroth-Wilding16,Pichotta:16,Mostafazadeh:17}. This line of work is related to story salads in that our aim of separating entangled narratives in a document mixture also leverages within-narrative coherence. Our work, however, is very different from narrative cloze: (i) we group sentences/events rather than predicting what happens next; (ii) crucially, the narrative coherence in story salads is \emph{in context}, in that a narrative clustering is only meaningful with respect to a particular document mixture (see Section 5, 6), while in narrative cloze the next event is predicted on a ``global'' level.\footnote{The story salad task is more similar to multichoice narrative cloze \cite{Granroth-Wilding16} in this regard, but formulated categorically differently.}

Working with labeled story salad examples, we draw inspiration from previous work on supervised clustering \cite{Bilenko:04,Finley:05}. We also take advantage of the recent success of deep learning in leveraging  a continuous semantic space \cite{Pennington:14,Kiros:15,Mekala:17,Wieting:17,Wieting:17b} for word/sentence/event encoding;  neural components for enhanced supervised clustering \cite{Bilenko:04}, in particular LSTMs \cite{Hochreiter:97,Dai:15}, CNNs \cite{Kim:14,Conneau:17}, and attention mechanisms \cite{Bahdanau:15,Hermann:15,Lin:17}. By exploring our ability to pick apart story salads with these state-of-the-art NLP modeling tools,
we attempt to (i) show the value of the story salad task as a new NLP task that warrants extensive research; (ii) understand the nature of the task and the challenges it sets forth for NLP research in general.

The task of picking apart story salads is related to the task of conversation disentanglement \cite{Elsner:08,Wang:09,Jiang:18}, which is a clustering task of dividing a transcript into a set of distinct conversations. While superficially similar to our Story Salad task, conversation disentanglement focuses on dialogues and has many types of metadata available, such as time stamps, discourse information, and chat handles. Existing systems draw heavily on this metadata. Another related task is the distinction of on-topic and off-topic documents~\cite{Bekkerman:2008}, which is defined in terms of topical relatedness. In comparison, the story salad task offers opportunities for more in-depth reasoning, as we show below. 

\section{Data}
\label{sec:data}
Natural story salads arise when multiple messy narratives exist to describe the same event or outcome. Often this is because each contribution to the explanation only addresses a small aspect of the larger picture. We can directly simulate the confusion this kind of discourse creates by taking multiple narratives, cutting them into small pieces, and mixing them together.

\textbf{Data generation}. Story salads are generated by combining content from source documents and randomizing the sentence order of the resulting mixture. In order to ensure appropriately sized salads, we require that each source document contain at least eight sentences. Furthermore, to avoid problematically large salads, we pull paragraphs from source documents one at a time until the eight sentence minimum is met. While this procedure can be used to mix any number of documents, we currently present mixtures of two documents.

We utilize two different corpora as sources for story salad generation: (i) the subset of New York Times articles presented within English Gigaword \cite{david2003english} and (ii) English Wikipedia\footnote{Wikipedia dump pulled on January 20, 2018.} \cite{wiki}. An overview of the datasets is available in Table \ref{tab:data-size}.

\begin{table}[!t]
\begin{center}
\scalebox{0.7}{
\begin{tabular}{lrrrr}
\toprule
Dataset & Salads & Total Words & $\mu$ Words/Salad & $\cos$ (test) \\
\midrule
\textsc{\s{nyt}} & 573,681 & 217,841,716 & 379.726 & 0.33 \\
\textsc{\s{nyt-hard}} & 1,000 & 20,149 & 438.220 & 0.56 \\
\textsc{\s{wiki}} & 500,000 & 197,175,135 & 394.350 & 0.35 \\
\textsc{\s{wiki-hard}} & 50,374 & 21,266,243 & 422.167 & 0.64 \\
\bottomrule
\end{tabular}}
\end{center}
\caption{Statistics of the datasets we present. The average topic cosine similarity scores ($\cos$) between the two narratives in document mixtures are computed from the test sets. The \textsc{nyt},  \textsc{wiki} and \textsc{wiki-hard} salads are divided into 80\%/20\% train/test splits, while the smaller \textsc{nyt-hard} is treated entirely as test.}
\label{tab:data-size}
\end{table}

\textbf{Gigaword}. From the New York Times subset of Gigaword, we compiled a set of 573,681 mixtures we call \s{nyt}. Each mixture in this set is constructed from source articles pulled from the same month and year. Because this temporal constraint is the only restriction put on what articles can be mixed, it is possible for a salad to be constructed from topically disparate source documents (e.g., a restaurant review and a political story). We intend \s{nyt} to be relatively easy on the whole as a result of this design choice.

However, it is also possible for articles about dominant news stories and trends (e.g., the OJ Simpson trial in the summer of 1994) to be mixed as a result of the same temporal constraint. We therefore pulled out a curated subset of \s{nyt} consisting only of salads generated from highly topically similar source documents which we call \s{nyt-hard}. This subset consists of the 1,000 salads where the source documents are most topically similar. We calculate topic similarity scores by computing the cosine similarity  between the average word embeddings for each source document (denoted $\cos$ hereafter)
\begin{equation}
\cos(d) = \frac{g(\omega_1)\cdot g(\omega_2)}{\parallel g(\omega_1)\parallel \parallel g(\omega_2)\parallel}
\end{equation}
where $\omega_1$ and $\omega_2$ are the source documents, $g$ is a function that computes the average word embedding of a document, and $d$ is the salad under evaluation. The $\cos$ scores on the test portion of the datasets are presented in Table \ref{tab:data-size}.

\textbf{Wikipedia}. From Wikipedia, we present an additional set of 500k salads constructed by combining random articles which we call \s{wiki}.

We also leverage Wikipedia category membership as a form of  human-annotated topic information. We use this to create a set of 50,374 salads, henceforth called \s{wiki-hard}, by restricting the domain of articles to only those appearing in categories containing the words \textit{conflict} and \textit{war}. Each mixture in this set is generated from source articles from the same category in order to produce highly difficult mixtures. We intend this to be a challenge set in this domain as the constituent articles for a given mixture are intentionally selected to be closely related. While we have used the category information to construct an intentionally very difficult set for this paper, we note that this procedure can be used to create sets of varying difficulty.

The fact that \s{wiki-hard} is generated from human-annotated category labels differentiates it from \s{nyt-hard} in the source of its difficulty. After manually reviewing 20 samples from each \s{*-hard} dataset, we found that \s{nyt-hard} more frequently contains salads that are impossible for humans to pick apart while \s{wiki-hard} more frequently contains salads that are possible, though challenging. In particular, in 9 out of 20 \s{wiki-hard} salads we found that access to world knowledge and inference would be beneficial. Nevertheless, the two \s{*-hard} datasets are both high in topic similarity (Table \ref{tab:data-size}).

In Figure~\ref{tab:wiki-hard-ex} we present sentences from a sample \s{wiki-hard} salad that can be solved with world knowledge. In this salad, we learn about two individuals. We can tell that Randle, born in 1855, is unlikely to also have been enrolled in high school in 1913 at the age of 58. We also learn that Randle was a doctor, while Martins, the other individual, was involved in theater. From this, we can deduce that the individual who ``also worked as a wrestler'' is more likely to be Martins than Randle.

\begin{figure}[t]
\begin{center}
\scalebox{0.70}{
\begin{tabular}{@{}p{28em}}
\toprule
\textcolor{NavyBlue}{(A) John K. Randle was born on 1 February 1855, son of Thomas Randle, a liberated slave from an Oyo village in the west of what is now Nigeria.} 
\textcolor{Sepia}{(B) In 1913 he was enrolled in Eko Boys High School but dropped out.}
\textcolor{Sepia}{(B) Martins joined the theatre and from there took on various theatre jobs to survive.}
\textcolor{NavyBlue}{(A) Born in Sierra Leone , he was one of the first West Africans to qualify as a doctor in the United Kingdom.} 
\textcolor{Sepia}{(B) He also worked as a wrestler (known as ``Black Butcher Johnson'').}\\
\bottomrule
\end{tabular}}
\end{center}
\caption{A story salad from \s{wiki-hard}, sourced from articles belonging to the \textit{Nigerian people of World War I} category. The sentences from this salad have been rearranged for clearer presentation.}
\label{tab:wiki-hard-ex}
\end{figure}

\textbf{Event Representation}. Finally, we explore a form of document representation that has been shown to be useful in narrative schema learning, a related task. We include variants of \s{nyt} and \s{nyt-hard} with story salads consisting of event tuple representations instead of natural language sentence representations, as in \citet{Pichotta:16}. We label these variants as \s{nyt-event} and \s{nyt-event-hard}. Event tuples are in the form $<$\s{VERB}, \s{SUBJ}, \s{DOBJ},  \s{PREP}, \s{POBJ}$>$, where as many preposition and prepositional object pairs as necessary are allowed.\footnote{Event tuples are extracted via the extractor presented in \citet{pximplicit}, and copular verbs are not treated as events, meaning that some sentences translate to null events.}

\textbf{Summary}. The story salads we present here are, in the end, simpler than those that occur naturally in the news or on social media: for one thing, sentences drawn from a document written by a single author should exhibit a high degree of coherence. We have also shown that we can use Wikipedia category annotations to produce large-scale story salad datasets with customizable levels of difficulty, enabling us to increase the difficulty of the task as performance increases. In the following section, we see that both our standard and \s{*-hard} mixtures are challenging for current models. Furthermore, our \s{wiki-hard} dataset contains salads featuring conflicting information and is an attractive setting for building models with deeper reasoning capabilities.

\section{Models}
\label{sec:models}

\begin{figure*}[!t]
\begin{center}
\includegraphics[width=150mm,trim=0 70mm 0mm 30mm]{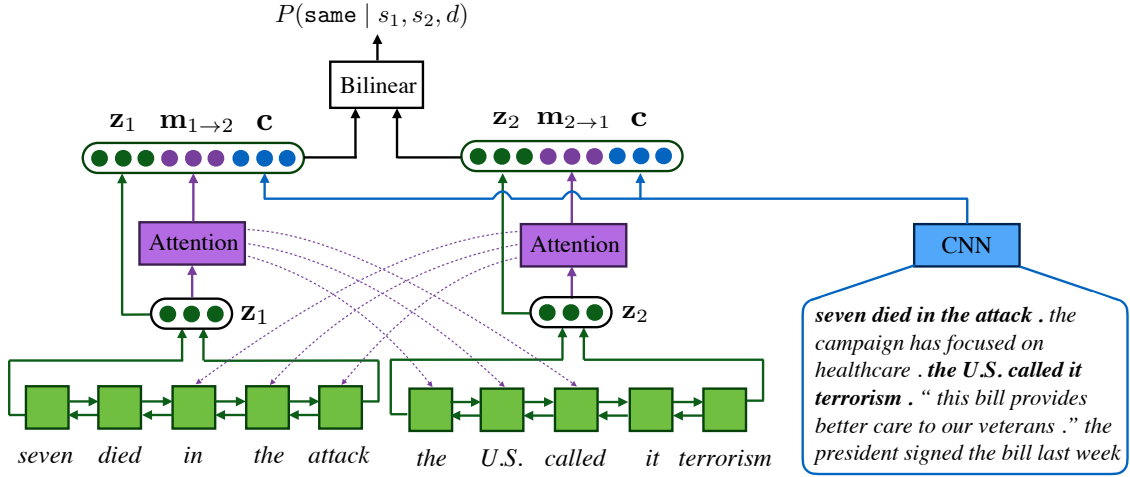}
\end{center}
\caption{BiLSTM sentence pair classifier to determine whether $s_1$ and $s_2$ are from the same narrative, augmented with a mutual attention and a context reader. The three subcomponents --- the BiLSTM, the mutual attention mechanism, and the context reader --- each produce vectors, denoted as $\bm{z}, \bm{m}, \bm{c}$ respectively. In the basic \s{bilstm} model, only $\bm{z}$ is fed to the bilinear layer (Eq. 2), while more sophisticated models incorporate the additional mutual attention and context vectors.}
\label{fig:bilstm-att-context}
\end{figure*}

We treat the story salad task as a narrative clustering task where, in our dataset, each salad is comprised of two clusters. Accordingly, the first baselines we consider are standard clustering approaches.

\textbf{Baselines}. Our first baseline is a simple \emph{uniform baseline} (hereafter \bs{unif}), where we assign all sentences in a document mixture to a single cluster. Under \textsc{unif} the clustering accuracy is the percentage of the majority-cluster sentences, e.g. if a mixture has 7 sentences from one narrative and 3 from the other, then the accuracy is 0.7. 

Additionally, we explore a family of baselines that consist of clustering off-the-shelf sentence embeddings. We choose k-medoids\footnote{K-medoids is chosen as a substitute for k-means because the latter does not extend easily to our classifier-aided neural models: it does not work when only pairwise distances are available. In empirical evaluation we found k-means and k-medoids to produce very similar accuracy scores when using off-the-shelf embeddings. Experiments with hierarchical agglomerative clustering (not reported here) showed it to perform worse than either method.} (hereafter \bs{km}) as our clustering algorithm. For sentence embeddings, we experimented with (i) averaged 300D GloVe embeddings \cite{Pennington:14}, which have been shown to produce surprisingly strong performance in a variety of text classification tasks \cite{Iyyer:15,Coates:18}; (ii) skip-thought embeddings \cite{Kiros:15}; and (iii) SCDV \cite{Mekala:17}, a multisense-aware sentence embedding algorithm which builds upon pretrained GloVe embeddings using a Gaussian mixture model. Averaged GloVe embeddings gave the best performance in our experiments; to avoid clutter, we only report those results henceforth.

\textbf{Neural supervised clustering}. Our baselines work directly on sentence embeddings and therefore ignore the task-specific supervision available in our labeled training data. Inspired by the work in \citet{Bilenko:04} and \citet{Finley:05} on supervised clustering, we aim to exploit this supervision using a learned distance metric in our clustering.\footnote{In early experiments, another strong candidate we tried is a joint model of a sentence autoencoder and a clustering algorithm \cite{Yang:17}. However, this produces subpar performance (weaker than the strongest baseline), due partially to scalability issues in learning these jointly.}

Figure~\ref{fig:bilstm-att-context} shows our model, which produces a distribution $P(\textrm{\texttt{same}} \mid s_1,s_2,d)$: the probability that two sentences $s_1$ and $s_2$ taken from document mixture $d$ are in the same cluster. We train this model as a binary classifier on sampled pairs of sentences to distinguish same-narrative sentence pairs (positive examples) from different-narrative pairs (negative examples). $1 - P(\textrm{\texttt{same}} \mid s_1,s_2,d)$ is then used by the clusterer as the pairwise distance metric. Given the pairwise distance between all sentence pairs in a mixture, the \textsc{km} algorithm can then be applied to cluster sentences into two narratives.

Our classifier is a neural network model built on top of LSTM sentence encoders, which perform well at similar text classification tasks \cite{Dai:15,Liu:16}.\footnote{Experiments with convolutional encoders here yielded somewhat worse results.} Denoting a sentence as the list of embeddings of its constituent words: $s = \{\bm{w}_1,\dots,\bm{w}_M\}$, we first encode it as a sentence embedding $\bm{z}$ with a bidirectional LSTM $\bm{z} = \texttt{BiLSTM}(s)$ and then 
compute the probability score with a bilinear layer:
\begin{equation}
P(\texttt{same}\mid s_1,s_2) = \sigma(\bm{z}_1^TW\bm{z}_2)
\end{equation}
This model corresponds to the green subset of Figure \ref{fig:bilstm-att-context}.

\textbf{Stronger models}. There are two additional effects we might want our model to capture. First, whether two sentences are from the same narrative cannot be determined globally: there aren't two ``globally-contrasted''\footnote{Two stories may be on the same topic and still form clearly different narratives. For example, both narratives in Figure \ref{tab-doc-mix} are regarding military conflict.} narratives (or bag-of-words based topics) from which sentences are sampled. In other words, sentences are always (pairwise) compared \emph{in the context} of the document mixture from which they are drawn. Second, we want to capture more in-depth interactions between sentences: our sentence embedding scheme for a sentence $s_1$ should exploit its point of comparison $s_2$ and encode $s_1$ with a view of similarities to and differences with $s_2$. This type of technique has been useful in tasks like natural language inference (NLI) \cite{Bowman:15,Peters:18}.

To improve contextualization, we add a CNN-based context encoder to the BiLSTM classifier: the reader embeds the whole document salad at hand into a vector.
Formally, we compute $\bm{c} = \textrm{CNN}(d)$, where in this case CNN denotes a single convolution layer with max pooling in the style of \newcite{Kim:14} and $d$ is the concatenation of all sentences in the mixture. This component is shown in blue in Figure~\ref{fig:bilstm-att-context}. The context vector $\bm{c}$ is then appended to $\bm{z}$ and fed into the bilinear layer.

To capture the interaction between two sentences in a pair, we employ a \emph{mutual attention} mechanism, which is similar to the attentive reader \cite{Hermann:15}. Let $\bm{e}_{i,1\ldots n}$ denote the BiLSTM outputs for the tokens of sentence $i$. Given the encoding $\bm{z}_1$ of sentence $s_1$, we compute attention weights and a representation of $s_2$ as follows:
\begin{align*}
\bm{\alpha}_{1 \rightarrow 2} = \textrm{softmax}_j (\bm{z}_1^\top \bm{e}_{2,j})\\
\bm{m}_{1\rightarrow 2} = \sum_{j} \alpha_{1 \rightarrow 2,j}\ \bm{e}_{2,j} 
\end{align*}
We compute $\bm{m}_{2\rightarrow 1}$ analogously. This process is shown in purple in Figure \ref{fig:bilstm-att-context}. The $\bm{m}$ vectors are used as additional inputs to the bilinear layer.

For comprehensive ablation, we experiment with four variants of neural classifiers: (i) BiLSTM alone (\bs{bilstm}); (ii) BiLSTM + mutual attention (\bs{bilstm-mt}); (iii) BiLSTM + context (\bs{bilstm-ctx}); and (iv) BiLSTM + mutual attention and context (\bs{bilstm-mt-ctx}).

\textbf{Event-based models}. For the event-based variants of the datasets, \s{nyt-event} and \s{nyt-event-hard}, we build three models: (i) \s{ffnn-bilstm}: we input a sentence as a sequence of event embeddings rather than word embeddings as in \s{bilstm}, where a feedforward layer maps the words in an event tuple to an event embedding; (ii) \s{ffnn-bilstm-mt-ctx}: replacing the base \s{bilstm} in (i) with our best model which is enhanced with mutual attention and contextualization; (iii) \s{ffnn-bilstm-mt-ctx-pretrain}: a variant of (ii) that is based on the event embedding pretraining method\footnote{In \citet{Weber:18}, a more complex tensor-based model is applied. Using exactly same method in our experiments we obtain weaker results.} described in \citet{Weber:18}, where events are encoded with a feedforward net (same as (i)) and trained with a word2Vec-like objective, encouraging events that co-occur in the same narrative to have more similar embeddings.


\section{Experiments and Analysis}
\label{sec:experiments-analysis}

\textbf{Experimental setup}. To stave off sparsity, we impose a vocabulary cut by using only the 100k most frequent lemmas. To evaluate on \s{nyt}, \s{nyt-event}, \s{wiki} and \s{wiki-hard}, we sample 20k unique salads (from their respective test portions\footnote{Test sets available with data release.}) to use for both the sentence and event versions of the experiments. For \s{wiki-hard}, the training combines the training portions of both \s{wiki} and \s{wiki-hard}. For \s{nyt-hard}, we train on the training portion of \s{nyt} and evaluate on \s{nyt-hard} in full as a test set. 

All the neural components are constructed with TensorFlow and use the same hyperparameters across variants: a 2-layer BiLSTM, learning rate 1e-5 with Adam \cite{Kingma:14d}, dropout \cite{Srivastava:14} rate 0.3 on all layers, and Xavier initialization \cite{Glorot:10}. To create training pairs for the neural classifiers, we randomly sample sentence pairs balanced between same-narrative and different-narrative pairs. We train with a batch size of 32 and stop when an epoch yields less than 0.001\% accuracy improvement on the validation set, which is 5\% of mixtures sampled from the training data beforehand (the models are not trained on the validation sample). For \s{km} we use the default configurations of off-the-shelf software.\footnote{ \burl{github.com/letiantian/kmedoids}}

\textbf{Evaluation}. We evaluate all models in terms of a \emph{clustering accuracy} metric (hereafter \textsc{ca}), which is a simple extension from the conventional accuracy metric: 
we calculate the ratio of correctly clustered sentences in a document mixture, averaged over test mixtures.
Given a document mixture $d_i$, we call its component documents $A$ and $B$. Let \texttt{pred} be a function that does the clustering by mapping each sentence $s_{i,n}$ of mixture $d_i$ to either A or B, and \texttt{true}$_{AB}$ a function that returns the original pseudo-labels (i.e. $\{A,B\}$) as they are, and \texttt{true}$_{BA}$ flips the pseudo-labels, i.e. $A\rightarrow B$ and $B\rightarrow A$. Then the clustering accuracy for document $d_i$ by \texttt{pred} is

{\small
\begin{align*}
\textsc{ca}(d_i, \texttt{pred}, \texttt{true}) = \texttt{max}\{&\textsc{c}(d_i,\texttt{pred},\texttt{true}_{AB}),\\
&\textsc{c}(d_i,\texttt{pred},\texttt{true}_{BA})\}\\
\s{c}(d_i,\texttt{pred},\texttt{true}) = \ \ \ \ \ \ \ \ \ \ & \\ \frac{1}{N_i}\sum_n\mathds{1}[\texttt{pred}(s_{i,n}&)=\texttt{true}(s_{i,n})]
\end{align*}}
where $s_{i,n}$ is the $n$-th sentence of mixture $d_i$.

\begin{table}[!t]
\begin{center}
\scalebox{0.77}{
\begin{tabular}{lcccc}
\toprule
Model & \s{nyt} & \s{wiki} & \s{nyt-hard} & \s{wiki-hard} \\
\midrule
\s{unif} & 52.7 & 50.9 & 52.5 &  51.2 \\
\s{km} & 76.4 & 74.9 & 59.8 & 60.4 \\
\midrule
\s{bilstm} & 78.5 & 76.2 & 55.3 & 59.8 \\
\s{bilstm-mt} & 80.8 & 78.8 & 56.7 & 61.3 \\
\s{bilstm-ctx} & 82.6 & 78.9 & 63.8 & 63.7 \\
\s{bilstm-mt-ctx} & \textbf{84.9} & \textbf{81.8} & \textbf{68.0} & \textbf{66.6} \\
\bottomrule
\end{tabular}}
\end{center}
\caption{Clustering accuracy (\s{ca}) results from the sentence based experiments. More sophisticated models do better across all datasets, particularly on \s{*-hard} tasks, which are substantially more challenging.}
\label{tab:sentence}
\end{table}

\textbf{Sentence based models}. First we evaluate the sentence based models. We first run the \s{unif} baseline on all our datasets, where we obtain near-50\% clustering accuracy. This indicates that the data are all balanced in the number of sentences in the two narratives of the mixtures. We then run k-medoids (\s{km}) on sentence embeddings as a baseline to compare to the classifier-aided models. The results are summarized in Table \ref{tab:sentence}.

We first observe that the \s{km} is a strong baseline and outperforms the supervised \s{bilstm} system in the harder settings. Adding the mutual attention mechanism and contextualization, however, improve \s{bilstm} substantially. In addition, the performance boost from the two components seems more or less orthogonal, as shown by the much stronger accuracy of the combined model (i.e. \s{bilstm-mt-ctx}) than the models with a single component (i.e. \s{bilstm-mt} and \s{bilstm-ctx}). Overall, the large margin of \s{km} and classifier-aided models above the \s{unif} baseline indicates that separating story salads is a valid task where generalizable patterns can be exploited by machine learning techniques. 

\begin{figure}[!t]
\begin{center}
\includegraphics[width=\linewidth]{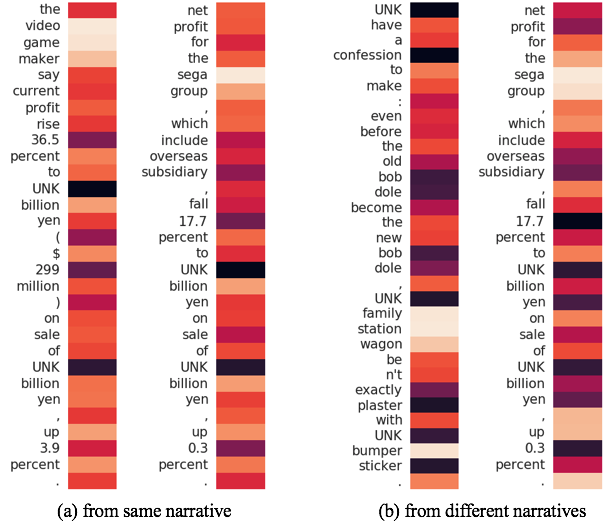}
\end{center}
\caption{Attention weight heatmaps for a random sample with \s{bilstm-mt-ctx}. Lighter color indicates higher attention weights. The two heatmaps in the same block are for the attention weights of one sentence attending to the other. In (a), we see related concepts being identified (\emph{video game} and \emph{sega}), while in (b), we see a contrast (\emph{family station wagon} and \emph{sega group}).}
\label{fig:mt-ctx-interaction}
\end{figure}

Why would the mutual attention mechanism help? Plotting the attention weights of randomly selected samples, we see distributionally similar words being attended to in Figure~\ref{fig:mt-ctx-interaction}a. Intuitively, a BiLSTM compresses a sentence into a single vector, leading to information loss \cite{Conneau:18}. Mutual attention enriches this representation by allowing us access to detailed information in sentences at word-level resolution by capturing lexical similarity. 
Even more interestingly, we observe a synergistic effect between mutual attention and contextualization: with the context reader added, we see high attention weights on words/phrases which bear little distributional similarity but are important for connecting/contrasting different narratives. For example, in Figure \ref{fig:mt-ctx-interaction}b, \emph{sega group} and \emph{family station wagon} are selected by the attention, despite not having similar words in the other sentences. These words are crucial in identifying the two narratives in this mixture: one is about a Japanese video game company, the other is on vehicle manufacturing in the U.S.  

Another observation is that all models see drastic reduction in accuracy in the \s{*-hard} version of the data. In fact, the clustering accuracy corresponds well with our topic similarity metric ($\cos$, Eq. 1; Table~\ref{tab:data-size}) across models. In addition, $\cos$ is negatively correlated with clustering accuracy for all mixtures (Table \ref{tab:corr}). 


\begin{table}[!t]
\begin{center}
\scalebox{0.80}{
\begin{tabular}{llcc}
\toprule
Type & Model & \s{nyt} & \s{wiki} \\
\midrule
\multirow{2}{*}{-\s{context}} & \s{bilstm} & $-0.40^*$ & $-0.43^*$ \\
& \s{bilstm-mt} & $-0.38^*$ & $-0.40^*$ \\
\midrule
\multirow{2}{*}{+\s{context}} & \s{bilstm-ctx} & $-0.31^*$ & $-0.30^*$ \\
& \s{bilstm-mt-ctx} & $-0.27^*$ & $-0.25^*$ \\
\bottomrule
\end{tabular}}
\end{center}
\caption{Spearman's $\rho$ correlation between clustering accuracy (\s{ca}) and topic similarity ($\cos$) in the evaluation with \s{nyt} and \s{wiki}. The p-values are all below 0.01 (indicated by *). Contextualized models (+\s{context}) are more robust to high topic similarity than their uncontextualized counterparts (-\s{context}), indicated by the lower negative correlation between their accuracy and topic similarity.}
\label{tab:corr}
\end{table}

From the results we also see that contextualization brings clear performance improvement. This supports our hypothesis that the Story Salad task is a nonstandard clustering task where the contrast of two narratives is only meaningful \emph{in the context} of the particular mixture where they reside, rather than on a corpus-general level. Taking the example in Figure 1, the Russian-Chechnya and the U.S.-Afghanistan narratives are contrasted in that mixture, but one can easily imagine a mixture where they are in the same narrative and are contrasted to another narrative on business affairs. Further, contextualized models are less vulnerable to the performance reduction on mixtures with high topic similarity: for one thing, contextualization improves performance over the base \s{bilstm} on both regular and 
\textsc{*-hard} datasets. Secondly, computing the correlation between clustering accuracy and topic similarity, we see a lower negative correlation for contextualized models, true for both \s{nyt} and \s{wiki} datasets (Table \ref{tab:corr}). 

\begin{table}[!t]
\begin{center}
\scalebox{0.75}{
\begin{tabular}{lcc}
\toprule
Model & \s{nyt-event} & \s{nyt-event-hard} \\
\midrule
\s{km} & 64.7 & 55.3 \\
\midrule
\s{ffnn-bilstm} & 64.9 & 54.8 \\
\s{*-mt-ctx} & 66.8 & 59.1 \\
\s{*-mt-ctx-pretrain} & \textbf{70.2} & \textbf{61.4} \\
\bottomrule
\end{tabular}}
\end{center}
\caption{Clustering accuracy (\s{ca}) results from the event based experiments. \s{*-mt-ctx} is a short hand for \s{ffnn-bilstm-mt-ctx}. The same notation applies for the following models.}
\label{tab:event}
\end{table}

\textbf{Event based models}. While the accuracy scores in the event based experiments are in general lower than those in the sentence based (Table \ref{tab:event}), overall we observe the same pattern that mutual attention and contextualization contribute substantially to the performance. More interestingly, the performance reduction on the topically highly similar \s{*-hard} is more mild compared to the sentence based experiments, which provides initial evidence that event-based narrative encoding allows the models to be more robust to distraction by lexical overlap in topically similar narratives. Finally we see that the event pretraining with \citet{Weber:18}'s technique brings additional improvement over a contextualized system. 

The results open up a door for future work: (i) our simple models do not make use of coreference, narrative schema or world knowledge, which are intuitively promising components to introduce (see, e.g., the salad in Figure 2); (ii) more sophisticated model architectures may help capture the information missed by our models: moving from the sentence version to the event version, we lose many words which may have provided crucial cues in the sentence-based experiments. 

\textbf{Error analysis}.  In order to understand the errors made by each model, we performed a manual analysis of a small sample of bad clusterings. In a sample of 60 mixtures from \s{nyt} (test set), we considered all clusterings for which accuracy was less than 0.65. Among the 60 mixtures, the base model had an accuracy this low for 27 mixtures, the \s{bilstm-mt} and \s{bilstm-ctx} model had 13 low-accuracy mixtures each, and \s{bilstm-mt-ctx} had 3. Each mixture was manually annotated by 2 annotators as being sourced from (i) thematically closely related documents (e.g., two stories on the same political event), (ii) thematically distinct documents (e.g., a political story and a sports story), or (iii) cannot tell.

Our analysis showed that the base \s{bilstm} model has difficulty even in cases where the source documents for the salad are thematically distinct. This was the case in 9 of 27 bad clusterings. The \s{bilstm-mt}, \s{bilstm-ctx} and \s{bilstm-mt-ctx} models not only have many fewer bad clusterings, they also show low accuracy almost exclusively in mixtures of related documents (2 cases of distinct documents for \s{bilstm-ctx}, none for \s{bilstm-mt} or \s{bilstm-mt-ctx}). Figure \ref{tab-bad-clust} shows an example of a bad clustering of two unrelated documents, produced by the base \s{bilstm} model.

\begin{figure}[t]
\begin{center}
\scalebox{0.70}{
\begin{tabular}{p{28em}}
\toprule
\textcolor{NavyBlue}{(A) lehman brothers be one of several investment bank eager to get UNK hand on state asset, across the nation and in massachusetts (A) former massachusetts governor william f. weld, a staunch supporter of privatization during UNK administration, have UNK in the hall of the state house, now as a corporate lawyer try to drum up support for the sale of lucrative state asset.} 
\textcolor{Sepia}{(B) officially, the rays option dukes, 22, to class a vero beach and place UNK on the temporary inactive list, where UNK will remain for an undetermined amount of time as UNK undergo counseling. (B) UNK apparently will receive UNK \$ 380,000 major-league salary .}\\
\bottomrule
\end{tabular}}
\end{center}
\caption{An example of (preprocessed) sentences from two unrelated documents being that have been clustered into a single cluster by the base model. \textcolor{NavyBlue}{Document (A)} is an article about proposed privatization of public assets, while \textcolor{Sepia}{Document (B)} is an article about happenings in Major League Baseball.}
\label{tab-bad-clust}
\end{figure}

In a second study, we rated the same 60 samples by their difficulty for a human,  focusing in particular on mixtures that went from low performance (0.5-0.65) in the \s{bilstm} model to high performance (0.8-1.0) in another model. For \s{bilstm-ctx} we find that only 2 out of 11 mixtures with such marked improvement over \s{bilstm} were hard for humans; for \s{bilstm-mt} only 1 out of 9 markedly improved mixtures was hard for humans. But for \s{bilstm-mt-ctx}, 8 out of 17 markedly improved mixtures were hard for humans, indicating that more sophisticated models do better not only on easy but also on hard cases. 

In a third small study, we compare \s{nyt-hard} and \s{wiki-hard} for their difficulty for humans, looking at 20 mixtures each. Here, very interestingly, we find more mixtures that are impossible for humans in \s{nyt-hard} (10 cases, example in Figure \ref{tab-nyt-hard-example}) than \s{wiki-hard} (3 cases). This presents a clear discrepancy between difficulty for humans and difficulty for models: the models do better on \s{nyt-hard} which is harder for us. While we would not want to draw strong conclusions from a small sample, this hints at possibilities of future work where world knowledge, which is likely to be orthogonal to the information picked up by the models, can be introduced to improve performance (e.g. \citet{Wang:18}).

Note that unlike many other NLP tasks where human performance sets the ceiling for the best achievable results (e.g. span-prediction based question answering \cite{Rajpurkar:16}, where all the information needed for the correct answer is available in the input), successfully picking apart narratives in a story salad may require consulting an external knowledge base, which affords machine learning models a clear advantage over humans. For example, recognizing that \emph{Commander Kamal} is likely to be Afghani based on his name, which is not knowledge every reader possesses, would allow us to successfully cluster the sentence with the U.S.-Afghanistan narrative rather than the Russian-Chechnya narrative. 

\begin{figure}[!t]
\begin{center}
\scalebox{0.70}{
\begin{tabular}{@{}p{28em}}
\toprule
\textcolor{NavyBlue}{(A) The most basic question face the country on energy be how to keep supply and demand in line. The Democrats would say : ``what can UNK do to make good use of what UNK have get?''}
\textcolor{Sepia}{(B) Oil price dominate the 31-minute news conference, hold here near pittsburgh.}
\textcolor{Sepia}{(B) Vice President Al Gore hold UNK first news conference in 67 day on Friday, defend UNK call for the release of oil from the government's stockpile and and vow that UNK would ``confront friend and foe alike'' over the marketing of violent entertainment to child, despite the million in donation UNK receive from Hollywood.}
\textcolor{NavyBlue}{(A) With oil price up, consumer agitate and the winter heating season loom, Vice President Al Gore and Gov. George W. Bush be go at UNK on energy policy, seek to draw sharp distinction over an issue on which both candidate have political vulnerability.}
\textcolor{Sepia}{(B) On other topic, Gore say UNK be not nervous about the upcoming debate, but be incredulous when a reporter ask whether UNK be confident UNK have the election lock up.}
\textcolor{NavyBlue}{(A) Bush, who criticize the decision as a political ploy to drive down price just ahead of election day, be schedule to discuss energy policy in a speech on Friday.}
\textcolor{Sepia}{(B) On Friday, Bush call Gore a ``flip-flopper'', say UNK proposal to tap into the reserve be a political ploy.} \\
\bottomrule
\end{tabular}}
\end{center}
\caption{Part of  a story salad that is impossible for a human to pick apart (source: \s{nyt-hard}). ``UNK'' represents out-of-vocabulary tokens, and all the words are lemmatized. Both narratives, i.e. \textcolor{NavyBlue}{(A)} and \textcolor{Sepia}{(B)} involve the characters Al Gore and George Bush, and both are on the topic of energy, with strongly overlapping vocabulary.}
\label{tab-nyt-hard-example}
\end{figure}


\section{Conclusion}
\label{sec:conclusion}

We have presented a technique to generate \emph{Story Salads}, mixtures of multiple narratives, at scale. We have demonstrated that the difficulty of these mixtures can be manipulated either based on document similarity or based on human-created document categories.  This data gives rise to a challenging binary clustering task (but easily extended to $n$-ary), where the aim is to group sentences that come from the same original narrative. As coherence plays an important role in this task, the task is related to work on narrative schemas~\cite{Chambers:08,Pichotta:16} and textual coherence~\cite{Barzilay:08,Logeswaran:18}. The automated and scalable data generation technique allows for the use of neural models, which need large amounts of training data.

Conducting a series of preliminary experiments on the data with common unsupervised clustering algorithms~\cite{Cao:10} and variants of neural network-based \cite{Kim:14,Dai:15,Liu:16} supervised clustering \cite{Bilenko:04,Finley:05} models, we have (i) verified the validity of the task where generalizable patterns can be learned through machine learning techniques; (ii) shown that this is a nonstandard clustering task in which the contrast between narratives is \emph{in context} as opposed to global; (iii) found that there is a class of mixtures that are doable for humans but very difficult for our current models, and that in particular the category-based method creates a high proportion of such mixtures.

Our work opens up a large number of directions for future research. First, while our models obtain strong results on simpler story salads, they have low performance on more difficult mixtures with high topical similarity. Second, there are many intuitively promising sources of information that we have not explored, such as coreference. And third, our models rely on pairwise similarity-based coherence learning, which leads to the natural question of whether structured prediction would improve performance.

\section*{Acknowledgments}

This research was supported by the DARPA AIDA program under AFRL grant FA8750-18-2-0017. Any opinions, findings, and conclusions or recommendations
expressed in this material are those of the authors and do not necessarily reflect the view of DARPA, DoD or the US government.  We  acknowledge  the  Texas  Advanced
Computing Center for providing grid resources that
contributed to these results. We are grateful to the anonymous reviewers for helpful
discussions.

\bibliography{emnlp2018}
\bibliographystyle{acl_natbib_nourl}

\end{document}